\documentclass[letterpaper, 10 pt, conference]{ieeeconf}  % Comment this line out if you need a4paper

\IEEEoverridecommandlockouts                              % This command is only needed if 
\usepackage{threeparttable}

\makeatletter
\newcommand\footnoteref[1]{\protected@xdef\@thefnmark{\ref{#1}}\@footnotemark}
\makeatother

\newcommand{\shorteq}{%
  \settowidth{\@tempdima}{-}% Width of hyphen
  \resizebox{\@tempdima}{\height}{=}%
}

\usepackage[colorlinks=true, citecolor=magenta]{hyperref}
\usepackage[table]{xcolor}

\usepackage{amssymb,fge}

\usepackage{amsmath, amssymb}
\usepackage{pifont}% http://ctan.org/pkg/pifont
\usepackage{subcaption}
\usepackage[utf8]{inputenc}
\usepackage[english]{babel}
\usepackage[linesnumbered,ruled,vlined]{algorithm2e}
\usepackage[font=small,belowskip=-1.5pt]{caption} % This causes problems while compiling to latex    
\usepackage{anyfontsize}

\usepackage{array}
\usepackage{graphicx}
\usepackage{amsfonts}
\usepackage[11pt]{moresize}
\usepackage{bm}
\usepackage{soul}
\usepackage{hhline}
\usepackage{multirow, makecell, multicol}
\usepackage{float}
\usepackage{booktabs,wrapfig}

\usepackage{amsthm}
\usepackage{color}
\usepackage{transparent}
\usepackage{url}
\usepackage{footmisc}
\usepackage{setspace}
\usepackage{mathtools}
\usepackage{gensymb}
\usepackage[export]{adjustbox}

\theoremstyle{plain}

\usepackage{graphicx}
\usepackage[T1]{fontenc}
\usepackage{pifont}
\usepackage{comment}
\usepackage{tabularx} % Required for tabularx environment
\usepackage{array} % Needed for defining new column type
\newcolumntype{Y}{>{\centering\arraybackslash}X}
\newcolumntype{Z}{>{\centering\arraybackslash}m{0.93cm}}

\newcolumntype{M}{>{\centering\arraybackslash}m{1.52cm}}

\title{\LARGE \bf
IDD-X: A Multi-View Dataset for Ego-relative Important Object Localization and Explanation in Dense and Unstructured Traffic
}

\author{Chirag Parikh$^{1}$, Rohit Saluja$^{2}$, C.V. Jawahar$^{1}$, Ravi Kiran Sarvadevabhatla$^{1}$ % <-this % stops a space
\thanks{$^{1}$Centre for Visual Information Technology, IIIT Hyderabad, India {\tt\small chirag.parikh@research.iiit.ac.in}}%
\thanks{$^{2}$Department of Computer Science, IIT Mandi, Mandi, India
        {\tt\small rohit@iitmandi.ac.in}}%
}

\begin{document}

\maketitle
\thispagestyle{empty}
\pagestyle{empty}

%%%%%%%%%%%%%%%%%%%%%%%%%%%%%%%%%%%%%%%%%%%%%%%%%%%%%%%%%%%%%%%%%%%%%%%%%%%%%%%%
\begin{abstract}

Intelligent vehicle systems require a deep understanding of the interplay between road conditions, surrounding entities, and the ego vehicle's driving behavior for safe and efficient navigation. This is particularly critical in developing countries where traffic situations are often dense and unstructured with heterogeneous road occupants. Existing datasets, predominantly geared towards structured and sparse traffic scenarios, fall short of capturing the complexity of driving in such environments. To fill this gap, we present IDD-X, a large-scale dual-view driving video dataset. With 697K bounding boxes, 9K important object tracks, and 1-12 objects per video, IDD-X offers comprehensive ego-relative annotations for multiple important road objects covering 10 categories and 19 explanation label categories. The dataset also incorporates rearview information to provide a more complete representation of the driving environment. We also introduce custom-designed deep networks aimed at multiple important object localization and per-object explanation prediction. Overall, our dataset and introduced prediction models form the foundation for studying how road conditions and surrounding entities affect driving behavior in complex traffic situations.
\end{abstract}

%%%%%%%%%%%%%%%%%%%%%%%%%%%%%%%%%%%%%%%%%%%%%%%%%%%%%%%%%%%%%%%%%%%%%%%%%%%%%%%%

\section{INTRODUCTION}

Understanding the influence of road and traffic conditions on ego vehicle's driving behavior is crucial for enabling explainability in automated driving decision-making. This capability helps in developing reliable and efficient intelligent vehicle systems. Unlike Western countries, developing nations contain dense and unstructured traffic situations, with heterogeneous road occupants (two-wheelers, animals, three-wheelers, etc), and static road objects (speed breakers, potholes, and traffic lights, etc.). Given the complex variety of road occupants and objects, it is important to know which of the multiple road entities influence the ego vehicle's driving behavior and more importantly, how they do so.

Existing datasets capture annotations for identifying important road entities \cite{li2022important}, \cite{gao2019goal}, \cite{malla2023drama}, with accompanying explanations in some cases \cite{xu2020explainable}, \cite{kim2018textual}, \cite{ramanishka2018toward}, \cite{malla2023drama}. In general, these datasets contain structured and sparsely populated traffic situations. In most cases, only a single road occupant is responsible for influencing the ego vehicle's driving behavior. On the contrary, in dense traffic scenarios found in developing countries, it is common to find multiple important road entities simultaneously affecting the ego vehicle's driving behavior. In structured driving scenarios, most vehicles adhere to traffic laws due to which there is limited variability in the ways they interact with the ego vehicle. However, road occupants usually do not obey the traffic rules in unstructured traffic scenarios. As a result, much more diverse variety of atypical and unexpected interaction patterns tend to exist. 

To ensure the representation of such scenarios among the datasets, we contribute a large-scale dual-view driving video dataset, IDD-X, captured in dense, heterogeneous, and unstructured traffic environments. The proposed dataset provides ego-relative annotations for multiple important road objects, their corresponding explanation labels, and the ego vehicle's driving behavior for the duration of the video clip. Drivers routinely utilize rearview information in their assessment of important objects. To account for this important aspect, IDD-X additionally captures road objects perceived in the rearview. By encompassing both front and rear views, IDD-X enables a more comprehensive analysis of driving behavior, providing a panoramic view of objects, their interactions, and the intricate cues that influence the driver's choices.

Harnessing the potential of our IDD-X dataset, we also introduce novel deep network architectures to address two key tasks - (1) multiple important object localization and (2) per object explanation prediction. These tasks serve as foundational components for unraveling the nuanced relationships between road conditions and ego vehicle's driving behavior in diverse and challenging traffic contexts.
% \subsection{Main Contributions}
Our contributions are summarised below.
\begin{enumerate}

\item We contribute IDD-X, the first ever driving video dataset with ego-relative annotations for \underline{both} multiple important object localization and corresponding explanations in dense and unstructured traffic. IDD-X contains 697K bounding boxes, 9K important object tracks, and 1-12 objects per video and comprehensive ego-relative annotations for multiple important road objects (10 categories) and explanations (19 categories).

\item In IDD-X, we provide the first dataset with \underline{multi-view} important object annotations for ego vehicle's driving behavior understanding.

% \item We introduce the first-ever dataset with multiple ego-vehicle driving behavior category annotations for understanding their relationship with important road objects.

\item We introduce custom-designed deep networks for a) multiple important object localization and b) per object explanation prediction with respect to the ego vehicle.
\end{enumerate}

% use multiplication
\begin{table*}[htbp]
  \caption{Comparison of IDD-X with public driving behavior understanding datasets. The I.O. abbreviations used in column names refer to the ego-relative Important Object. The column "Irregular Road Surface Description" refers to speed breakers and potholes as important object annotations.}
  \label{datasets_comparison}
  % \centering
  \begin{center}
  \scriptsize
  \hyphenpenalty=10000 % Discourage hyphenation
  \exhyphenpenalty=10000 % Discourage hyphenation
  \renewcommand{\arraystretch}{1.5}
  \begin{tabularx}{\textwidth}{cYYYYYYYYYYYY}
  % \begin{tabularx}{\textwidth}{ccZZZZZYZZZZ}
    \hline
    \textbf{Dataset} & \textbf{Multi-View} & \textbf{I.O. Location} & \textbf{I.O. Track} & \textbf{Irregular Road Surface Description} & \textbf{Dense Traffic} & \textbf{Unstructured Traffic} & \textbf{\#I.O. Bounding Boxes} & \textbf{\#I.O.s Per Video} & \textbf{\#I.O. Explanation Categories} & \textbf{\#I.O. Categories} & \textbf{\#I.O. Explanation Type} \\[1.5ex] \hline
    \textbf{IDD-X} & \ding{52} & \ding{52} & \ding{52} (\textbf{9K}) & \ding{52} & \ding{52} & \ding{52} & \textbf{697K} & \textbf{1-12} & \textbf{19} & \textbf{10} & Categorical \\
    DRAMA \cite{malla2023drama} & - & \ding{51} &-&-&-&-& 17K & 1 & - & 3 & Textual \\
    H3D \cite{li2022important} &-& \ding{51} &-&-&-&-& 8K & 1-3 & - & 4 & - \\
    BDD-OIA \cite{xu2020explainable} &-&-&-&-& \ding{51} &-& - & - & 8\footnotemark & 6 & Categorical \\
    OIE \cite{gao2019goal} &-& \ding{51} &-&-&-&-& 4K & 1-2 & - & 2 & - \\
    HDD \cite{ramanishka2018toward} &-&-&-&-&-&-& - & 1 & 5 & 4 & Categorical \\
    BDD-X \cite{kim2018textual} &-&-&-&-&-&-& - & - & - & - & Textual \\
    METEOR \cite{chandra2023meteor} &-&-&-&-& \ding{51} & \ding{51} & - & - & - & - & - \\
    \hline
  \end{tabularx}
  \end{center}
\end{table*}

Visit our project page \url{https://idd-x.github.io} for details.

\section{RELATED WORKS}

\textbf{Important Object Identification}: Identifying important road entities that influence the ego vehicle's driving behavior forms an important component of the generated behavior explanations. Approaches for important object identification use the ego vehicle's trajectory, the object's trajectory, and the video context information \cite{gao2019goal} \cite{li2022important}. However, these methods do not consider the category of road objects. Given the heterogeneous road object categories in our dataset, we explicitly incorporate object class data along with trajectory information for predicting its importance.
% This enables our model's generalization capability for heterogeneous traffic settings.
% 1. Point the existing approaches for IOId, and how is ours different:
%     a. Consideration of tracks (relative trajectory) for ID
%     b. Consideration of heterogeneous road object categories for improving the ID task.
% certain road object categories behave differently than others when they influence the ego's driving decisions.
 
 % for the ego's driving behavior 
 % applications like autonomous driving

\textbf{Important Object Explanation}: Explaining the ego vehicle's driving behavior in the light of identified important objects is crucial for reliable usage of vision-based driving models \cite{bojarski2016end} \cite{kim2017interpretable} \cite{xu2017end} \cite{wu2019end}. However, existing approaches \cite{kim2018textual} \cite{ben2022driving} \cite{jing2022inaction} \cite{li2020learning} do not identify important objects when generating explanations. Specifically, they do not localize road entities that form the basis for the explanation. This introduces ambiguity in the evaluation of generated explanations. The model proposed by Kim et al. \cite{kim2018textual} and Jing et al. \cite{jing2022inaction} provide pixel-level saliency maps along with explanations. However, these maps do not overlap reliably with causal road entities. Recently proposed approaches \cite{malla2023drama} \cite{xu2020explainable} provide object-level explanations using an RGB image or its optical flow image as input. However, these image-based methods do not utilize temporal information or trajectory of important objects for explanation prediction. In our proposed approach, we extract spatiotemporal features of important object tracks and utilize them for generating ego vehicle explanations.

\textbf{Datasets}: Multiple datasets exist for identifying risk objects~\cite{malla2023drama}, anomalous objects~\cite{yao2020and}, causal objects~\cite{ramanishka2018toward}, and important objects~\cite{gao2019goal,li2022important} that affect the ego vehicle’s driving behavior. However, the traffic scenarios in these datasets are typically sparse and mostly contain a single important object’s annotation in a driving sequence. Such datasets cannot be used to explain the dense traffic conditions where multiple road objects simultaneously influence the ego’s driving decisions. Although some datasets specifically capture dense traffic scenes \cite{xu2020explainable} \cite{chandra2023meteor} \cite{dokania2023idd}, they do not provide category and location information for ego-relative important objects. In contrast, our IDD-X dataset has spatial and temporal location annotations for \emph{multiple} important road objects that explain the ego’s driving situation in dense traffic. 
% Our dataset contains explanations for challenging unstructured traffic situations containing complex and unpredictable interaction patterns.

% Overall, 
% consideration of rear view
% provision of tracks for important objects
% representation of dense and unstructured traffic.
% Less dataset requirement as compared to training language-based models.

Our dataset offers explanations for challenging unstructured traffic situations containing complex and unpredictable interaction patterns which are often rare in structured driving scenarios found in existing datasets. Additionally, IDD-X is the first dataset to consider rearview information for important object annotations -- see Table \ref{datasets_comparison} for a detailed comparison. 
% Also, since we provide categorial explanations Less dataset requirement as compared to training language-based explanation models.
% To the best of our knowledge, 

\footnotetext{For a fair comparison with our dataset, explanations referring to at least one important or action-inducing object are considered and similar explanations for left and right turns are merged into one category.}

\begin{figure}
  \centering
  \includegraphics[width=\linewidth]{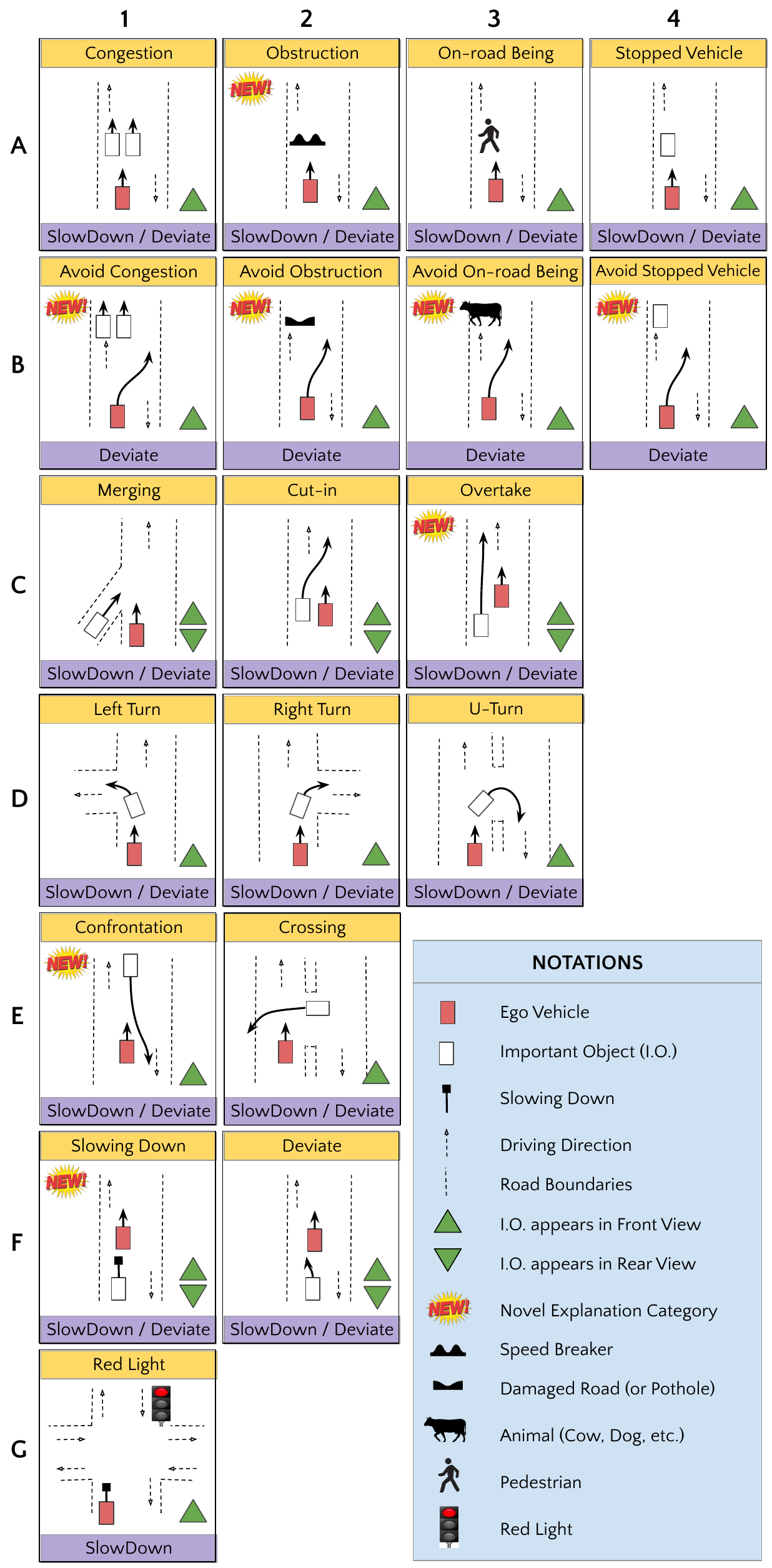}
  \caption{Bird Eye View illustration of Explanations for Important Objects in different traffic situations captured in the IDD-X dataset. Each row (A to G) and column (1 to 4) pair is an explanation card. Inside each card, the yellow header corresponds to the explanation category while the purple footer corresponds to the possible ego vehicle's driving behaviors in the demonstrated traffic situation. Icon notations used in the figure are defined in the bottom right corner. Important Object(s) in each card can either be visible in the front-camera view, in the rear-camera view, or in both of them. The corresponding notation for the views is shown in the bottom right corner of each card. The Driving Direction notation shows that the vehicles should ideally drive on their left-hand side of the road according to Indian traffic rules. The Novel Explanation Category icon means that the category is unique to this dataset, and has not been found in the existing driving datasets.}
  \label{fig:IDDX_Explanations_BEV}
\end{figure}

% The object's explanation is given with its category card number from Figure~\ref{fig:IDDX_Explanations_BEV} for reference and comparison of the maneuvering styles and the traffic situations.

\begin{figure*}
  \centering
  \includegraphics[width=0.918234\textwidth]{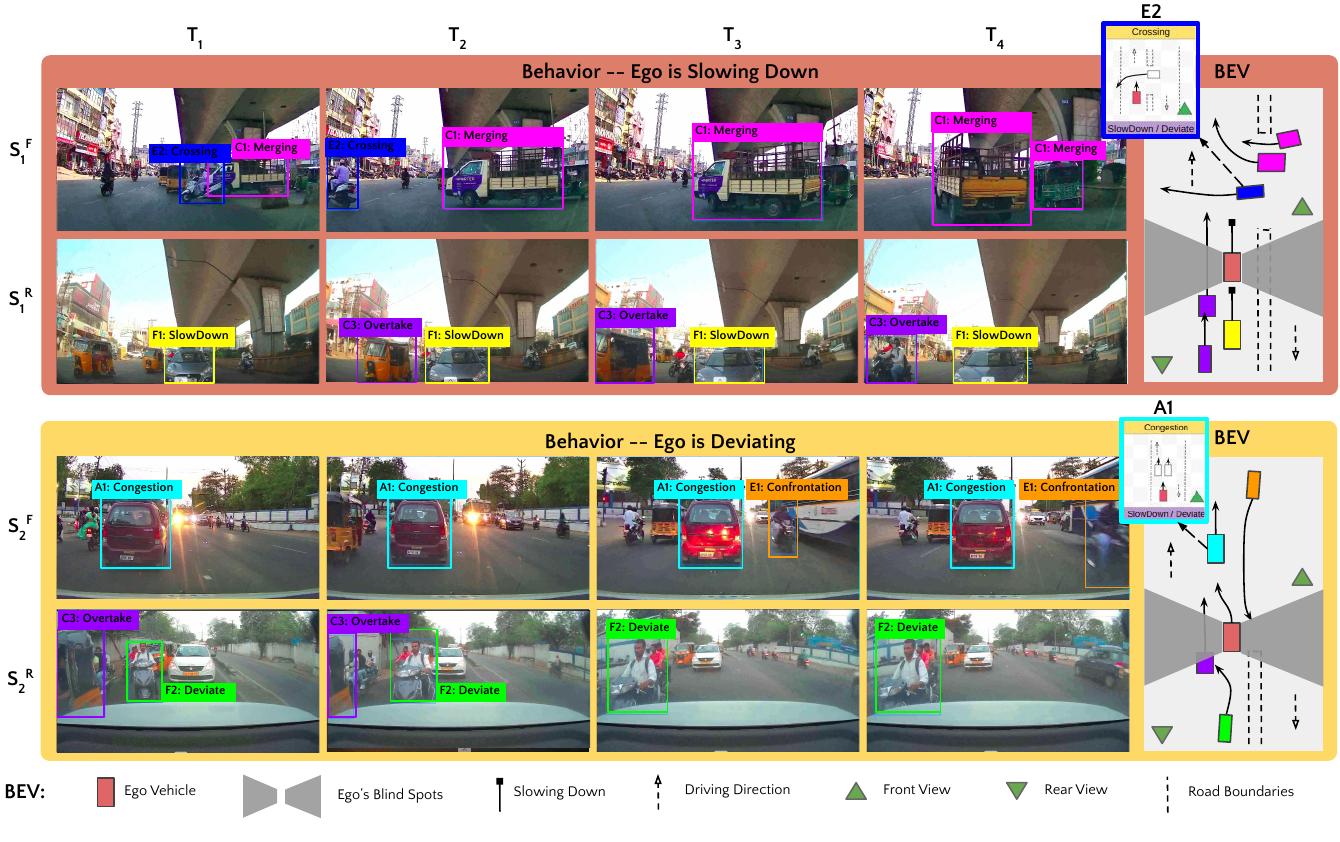}
  \caption{Samples of the annotated driving scenarios in IDD-X. The ego vehicle's driving behavior annotation is displayed at the top of each scenario (\( \text{S}_{1} \) and \( \text{S}_{2} \)). The important object location annotations in front and rear views (\( \text{S}_{1}^{\text{F}} \), \( \text{S}_{2}^{\text{F}} \), \( \text{S}_{1}^{\text{R}} \), \( \text{S}_{2}^{\text{R}} \)) at timeframes (\( \text{T}_{1} \) to \( \text{T}_{4} \)) are shown with colored bounding boxes. Unique colors are assigned to the objects on the basis of their explanation category. The object's explanation annotation is attached to each bounding box with the explanation card ID referenced from Figure~\ref{fig:IDDX_Explanations_BEV}. The Bird's Eye View (BEV) illustration shows the trajectory of ego vehicle and annotated important objects as observed in four timeframes. The icon notations used in BEV are defined at the bottom of this figure. Sample explanation cards are shown in BEV for comparison of the important object's maneuvering styles and traffic situations.}
  \label{fig:DatasetAnnotationSamples}
\end{figure*}
% For every explanation label in this figure, a unique color is assigned which is used to represent the important objects in the BEV. 
% comparison of the maneuvering patterns and the traffic situations

\section{THE DATASET}

% Navigating the diverse and chaotic road conditions in developing countries requires an understanding of the critical roles that the on-road traffic participants play in influencing the ego’s driving decision. To facilitate the advancement of intelligent vehicle systems tailored for these complex environments, we introduce IDD-X, a large-scale multi-view driving video dataset captured in dense, heterogeneous, and unstructured traffic situations. 

Our dataset IDD-X is specifically designed to explain ego vehicle’s driving behavior due to multiple important road objects. These objects affect the ego vehicle's driving decision and it is important to know the identity and role of the respective objects. To this end, we utilize front and rearview driving videos captured from vehicle-mounted cameras to assess the ego's surrounding traffic situation. Temporal modeling of road objects is crucial for describing complex and unpredictable driving maneuvers commonly observed in unstructured traffic. Therefore, individual object-level tracks are included. Our dataset also provides annotations for 10 different road object categories: car, motorcycle, autorickshaw, truck, bus, bicycle, person/animal, speed breaker and damaged road (pothole). If an object does not fall into any of the listed categories, it is assigned to `others' class. 

% Define an important object ()

% are given in the form of rationales
% Our dataset primarily focuses on explaining the ego’s driving behaviors or actions that are taken in response to external stimuli (i.e., road objects). 

Our dataset considers the following predominant driving situations: a) slowing down on straight roads, b) deviating on straight roads, and c) slowing down on left/right/U-turns. Every important object is associated with an ego-relative explanation during the driving situation. 
The explanations in IDD-X comprise a set of pre-defined categories that encompass the diversity of complex ego-relative interaction patterns of the heterogeneous road objects in unstructured traffic conditions. We define 19 different explanation categories for the important objects: congestion, obstruction, on-road living being, stopped vehicle, avoid congestion, avoid obstruction, avoid on-road living being, avoid stopped vehicle, merging, cut-in, overtake, confrontation, crossing, slow down, deviate, left turn, right turn, u-turn, red light. 

A bird's eye view illustration for each of these explanations in different traffic situations is shown in Figure~\ref{fig:IDDX_Explanations_BEV}. Each row in the figure corresponds to a group of explanations with similar interaction patterns between the important road objects and the ego vehicle. 
\begin{itemize}
\item Row A: Important objects appearing in the front of ego vehicle are in close interaction such that the ego vehicle's driving path is blocked. 
\item Row B: The objects are relatively far such that sufficient driving space is available for the ego vehicle to avoid or move ahead of the object by deviating from its path. 
\item Row C: Lateral interactions between the ego vehicle and the important road objects. 
\item Row D: Important objects doing a turning maneuver in front of the ego vehicle. 
\item Row E: Important objects approaching the ego vehicle such that the ego's future trajectory is hindered. 
\item Row F: Primitive important object explanations whose consideration is crucial for the ego vehicle to make safe and careful driving decisions. 
\end{itemize}

Broadly, Rows A, B, and G describe the passive influence of different road objects on ego vehicle's driving decision. Rows C, D, E, and F define the ego-relative maneuvering styles of the important road objects. Note that the situations in Figure~\ref{fig:IDDX_Explanations_BEV} are representative and variants may exist in data. For instance, in Row B, the ego vehicle may avoid the important objects from its left depending on the available driving space.

%Similarly, the lateral interactions in Row C may take place from either side (left/right) of the ego vehicle. Also, the road conditions may be different when these interactions take place. The Red Traffic Light in Row G need not necessarily be in a 4-way intersection or the speed breaker in Row A Column 2 may either be covering the entire or just the partial road width. 
% The same is true for the interaction patterns in Rows E and F. 
% To showcase the complexity of dense, heterogeneous, and unstructured traffic situations in our dataset,
% making it simpler (everything need not be described
% give a sample card for second example also and also give the explanations in a logical order.
Samples of annotated driving scenarios from our dataset are shown in Figure~\ref{fig:DatasetAnnotationSamples}. In scenario \(\text{S}_{1} \), the ego vehicle slows down because multiple important road objects (motorcycle, truck, and autorickshaw appearing in front view \( \text{S}_{1}^{\text{F}} \)) block ego vehicle's driving path. This shows that ego vehicle's driving behavior is influenced by the interactions of multiple road objects. Such scenarios are common in dense and unpredictable traffic conditions captured in our dataset. %In contrast, existing datasets capturing sparse traffic conditions mostly contain single important object annotation per driving scenario. 

The road objects appearing in the rearview may also impact the ego vehicle's driving decision. In scenario \(\text{S}_{1} \), the ego vehicle may collide with important objects (autorickshaw and motorcycle in rear view \( \text{S}_{1}^{\text{R}} \)) if ego chooses to deviate from its driving path. The ego vehicle may collide with important object (car in rear view) if it chooses to abruptly slow down. Therefore, both the front and rear views are crucial for comprehensive analysis of the ego vehicle's driving decisions.

% The diversity of complex and unpredictable interaction patterns by heterogeneous road occupants shown in Figure~\ref{fig:DatasetAnnotationSamples} represents the unstructured nature of the traffic in the IDD-X dataset. 

% In the following sections, we outline the data collection procedure, including details about the collection platform, sensor configurations, raw data details, and annotation procedure in Section~\ref{sec:data_creation}. Next, we provide a statistical analysis of the dataset in Section~\ref{sec:data_stats}.

% \begin{figure}
%   \centering
%   \includegraphics[width=\linewidth]{Annotation_Interface.pdf}
%   \caption{Annotation Interface. We use the open-source CVAT annotation tool for annotating the important object track, explanation class, object category, and the ego's driving behavior.}
%   \label{fig:AnnotationStrategy}
% \end{figure}

\subsection{Dataset Creation}
\label{sec:data_creation}

We recorded 85 hours of dual-view driving videos from urban, rural, and highway roads in and around Hyderabad city, India. The dataset was captured in day, night, rainy, cloudy, and sunny climates and weather conditions. The front-view and the rear-view videos were captured with camera resolutions of 2560 × 1440, and 1920 x 1080, recording at 25 fps. 

We used the open-source CVAT\footnote{
\href{https://github.com/openvinotoolkit/cvat}{\textcolor{blue}{https://github.com/openvinotoolkit/cvat}}} tool for annotating the IDD-X dataset. %The web interface of the tool is shown in Figure~\ref{fig:AnnotationStrategy}. 
The annotations are designed to capture all the important interactions of the surrounding road occupants when the ego vehicle takes a driving decision. 

\textit{Annotaion procedure:} The annotators watch the driving videos and filter out video intervals or scenarios where the ego driver either slows down or deviates from the influence of any road entity. These driving behaviors are called stimulus-driven actions which were initially defined in the HDD dataset \cite{ramanishka2018toward}. A total of 3635 driving scenarios were filtered out with their durations ranging from 0.3 to 11 seconds. These video intervals with annotated driving behaviors were analyzed by a different group of annotators (experienced drivers) to identify all the important road objects while imagining themselves driving the ego vehicle. The annotators create bounding boxes around such objects and track them by adjusting the bounding boxes across all frames at the frequency of 25Hz for the entire duration of the driving scenario. Next, they re-watch the entire video sequence, to specifically observe the interaction patterns of the important objects and assign suitable explanation labels from the pre-defined set of explanations described before. The annotators refer to the bird's eye view illustration similar to Figure~\ref{fig:IDDX_Explanations_BEV} for identifying the best-fitting explanation category for the road object. The explanation label and the object's category information are provided as the object's track attributes.

\subsection{Data Statistics}
\label{sec:data_stats}

\begin{figure}
  \centering
  \includegraphics[width=.93\linewidth]{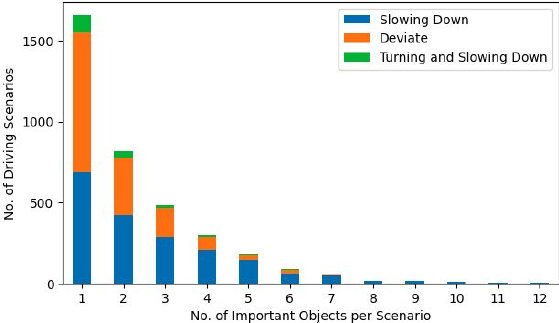}
  \caption{Statistics of the number of Important Objects in a driving scenario for every driving action class.}
  \label{fig:DatasetStatisticPlots1}
\end{figure}

\begin{figure}
  \centering
  \includegraphics[width=.93\linewidth]{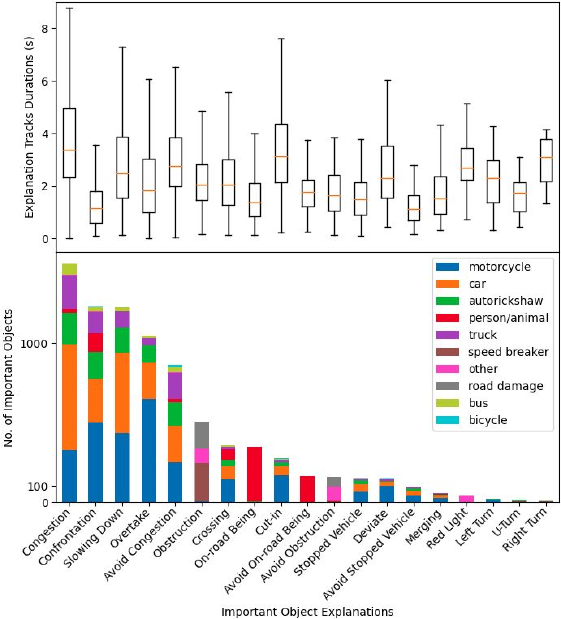}
  \caption{Distribution of Explanations for Important Objects for every road object category and the average duration of the corresponding object tracks.}
  \label{fig:DatasetStatisticPlots}
\end{figure}

 % with 1909 cases of slowing down on straight roads, 1544 cases of deviating on straight roads, and 182 cases of slowing down on turns
The dataset includes 3635 driving action scenarios containing 8689 important object annotations out of which 6427 appear in the front view and 2262 objects appear in the rear view. Figure~\ref{fig:DatasetStatisticPlots1} shows the distribution of the number of important objects observed per driving scenario. The larger number of observed important objects indicates denser traffic scenarios where 54\% of the total scenarios in the dataset constitute more than one important object per scene. The distribution of explanations for heterogeneous road objects is shown in Figure~\ref{fig:DatasetStatisticPlots}. A heavy-tailed distribution of the labels can be observed in the figure.
% durations of the important object tracks containing such explanations are also displayed in the same figure

% A total of 1978 driving scenarios contain multiple important object annotations while 1657 of them contain a single important object.

% \begin{figure*}
%   \includegraphics[width=\textwidth]{IDDX_model.pdf}
%   \caption{The proposed approaches for Important Object Localization and Explanation in Dense and Unstructured Traffic}
%   \label{figg:ModelArchitecture}
% \end{figure*}

\section{TASKS}

%Our goal is to provide a comprehensive understanding of the ego’s driving situation when the vehicle is under the influence of external stimuli. To accomplish the same, we consider the following tasks a) Important Object Localization for identifying all the critical objects in the driving scenario (Sec.~\ref{ssec:IOL}) and b) Important Object Explanation Prediction (Sec.~\ref{ssec:IOE}) for classifying the relationship of the critical object with respect to the ego vehicle. 

Given the inter-related nature of Important Object Localization and Important Object Explanation Prediction, we introduce a custom-designed multi-task deep network which obtains the related predictions for a given driving video. Figure~\ref{figg:ModelArchitecture} illustrates the deep network along with essential components and procedures for the aforementioned tasks. 

%%% In this section, we delineate the steps taken and the approaches employed to develop deep network architectures for the tasks mentioned.

\subsection{Important Object Localization}
\label{ssec:IOL}

In this task, we localize the important objects in a driving situation using their relative motion information. The approach for the task is divided into two steps 1) Multi-Object Tracking for localizing all the road objects observed in the driving scenario, and 2) Important Object Track Identification for predicting the importance of each object track. %Figure~\ref{figg:ModelArchitecture} shows the Important Object Selector Module with input RGB frames sequence and output important object category and track bounding boxes.

 % Per-class accuracy with overall accuracy is reported.
\begin{table}[!t]
    \caption{Driving Behavior Recognition. All scores are in \%.}
    \label{ar_results}
    % \centering
    \begin{center}
    \scriptsize
    \renewcommand{\arraystretch}{1.3}
    % \begin{tabularx}{\linewidth}{YYYYY}
    \begin{tabularx}{\linewidth}{ccccc}
    \hline
        \textbf{Modality} & \textbf{Slowdown} & \textbf{Deviate} & \textbf{Turn and Slowdown} & \textbf{Accuracy} \\ \hline
        RGB & 79.4 & 57.3 & 12 & 65.5 \\ %\hline
        \textbf{Flow} & \textbf{81.3} & \textbf{68.7} & \textbf{36} & \textbf{72.8} \\ \hline
    \end{tabularx}
    \end{center}
\end{table}

\textbf{Multi-Object Tracking}: For this sub-task, we first detect all the objects in a driving scene observed from the front view camera and then use a tracking algorithm to track all the detected objects in subsequent video frames. A YOLOv4 detection model~\cite{bochkovskiy2020yolov4} was applied over all the frames in the front-view driving scenarios to get bounding boxes of all the road occupants. These bounding boxes were provided as input to the SORT tracker~\cite{bewley2016simple} algorithm for their association across frames to get their unique track-ids. This gives us the tracks of all the objects in the driving scene with their class labels. 

\begin{figure}[!ht]
  \centering
  \includegraphics[width=\linewidth]{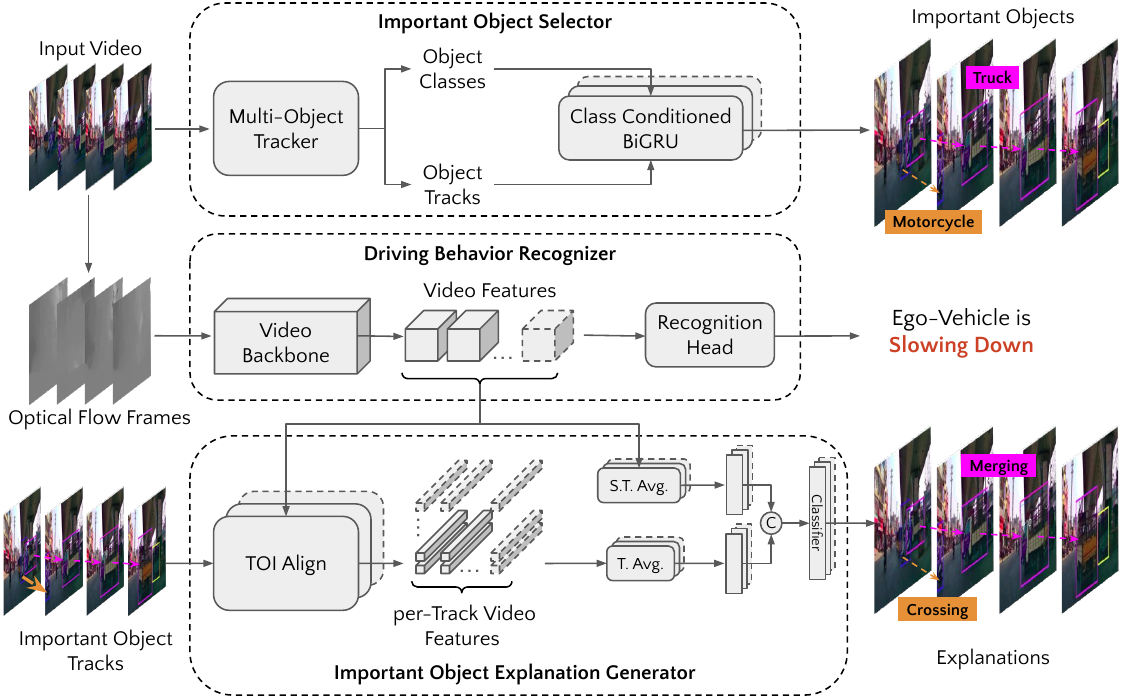}
  \caption{The proposed approaches for Important Object Localization and Explanation in Dense and Unstructured Traffic.}
  \label{figg:ModelArchitecture}
\end{figure}

\textbf{Important Object Track Identification}: This sub-task is formulated as a binary classification problem to predict the importance of an object track obtained from the tracking algorithm. We first prepare the training data with the binary importance labels for all the road objects in the driving scenes. Our dataset contains the ground-truth track information for the important road objects. We use this to label all the predicted object tracks, obtained from the previous sub-task, with the importance class. This is done by matching the predicted tracks with the ground truth tracks using spatial Intersection Over Union (IOU) between their bounding boxes averaged across all frames. The predicted object track is labeled as important if the IOU value is above a threshold otherwise it is marked as unimportant. This labeled training data is now used for importance prediction as described below.

The predicted object track's bounding boxes and its object category are provided as input to an object class conditioned Bi-directional Gated Recurrent Unit (BiGRU) model~\cite{650093} to output the importance class. A BiGRU model helps in effective temporal modeling of the object’s relative trajectory with respect to the ego vehicle. Additionally, it also considers the object category for differentiating between the bounding box size variations of different road objects. This is achieved by providing the encoded information of the object class along with its bounding box track coordinates at every time step inside the BiGRU model. Overall, our custom-designed approach helps in maintaining the relevant spatiotemporal context information throughout the track length. As shown in Figure~\ref{figg:ModelArchitecture}, this model is separately applied over all the object tracks for their importance prediction.

% Precision, Recall, and F1-Score are reported for the Important class.
\begin{table}[!ht]
    \caption{Important Object Track Identification. All scores in \%.}
    \label{IOI_results}
    % \centering
    \begin{center}
    \scriptsize
    \renewcommand{\arraystretch}{1.3}
    % \footnotesize
    % \begin{tabularx}{\linewidth}{YYYYY}
    % \begin{tabularx}{\linewidth}{BBBB}
    \begin{tabularx}{\linewidth}{ccYYY}
    \hline
        \textbf{Model} & \textbf{Conditioning} & \textbf{Precision} & \textbf{Recall} & \textbf{F1-Score} \\ \hline
        BiGRU & - & 35.0 & 86.1 & 49.8 \\%[0.7ex] %\hline
        Class Conditioned BiGRU & At t = 0 & 34.8 & 86.5 & 49.6 \\%[3.2ex] %\hline
        Class Conditioned BiGRU & At t = T & 34.9 & 85.6 & 49.6 \\%[3.2ex] %\hline
        \textbf{Class Conditioned BiGRU} & \textbf{At all t} & \textbf{35.8} & \textbf{86.8} & \textbf{50.7} \\ \hline
    \end{tabularx}
    \end{center}
\end{table}

\subsection{Important Object Explanation}
\label{ssec:IOE}

%The definition of the explanation labels in this dataset is based on the diverse interaction patterns observed in driving videos from the ego's point of view. To efficiently model such data, robust video features are required that can differentiate between the intricate motion patterns. To extract such features, we train a video recognition model which also fulfills an additional sub-task discussed below.

\begin{table*}[!ht]
    \caption{Important Object Explanation Prediction. F1-score for each explanation category is reported. All scores are in \%.}
    \label{explanation_results}
    % \centering
    \begin{center}
    \scriptsize
    \renewcommand{\arraystretch}{1.3}
    \hyphenpenalty=10000 % Discourage hyphenation
    \exhyphenpenalty=10000 % Discourage hyphenation
    % \begin{tabularx}{\textwidth}{YYYYYYYYYYYY}
    \begin{tabularx}{\textwidth}{c|YcYYYYYY|YY}
    \hline
        \textbf{Input Features} & \textbf{Congestion} & \textbf{Confrontation} & \textbf{Avoid Congestion} & \textbf{Overtaking} & \textbf{Crossing} & \textbf{Interfering Being} & \textbf{Cut-in} & \textbf{Avoid On-road Being} & \textbf{Avg. F1-Score} & \textbf{Wt. Avg. F1-Score} \\ \hline
        TOI-Aligned & \textbf{72.2} & 48.6 & \textbf{53.3} & \textbf{53.9} & 37.5 & 17.4 & 7.3 & 0.0 & 36.3 & 54.8 \\ %\hline
        \textbf{TOI-Aligned + Context} & 71.0 & \textbf{53.6} & 47.2 & 48.6 & \textbf{44.7} & \textbf{22.4} & \textbf{34.7} & \textbf{25.0} & \textbf{43.4} & \textbf{55.8} \\ \hline
    \end{tabularx}
    \end{center}
\end{table*}

\textbf{Driving Behavior Recognition}: We adopt the TSN model \cite{wang2018temporal} with Resnet-101 backbone for recognizing the driving behaviors in our dataset as it has demonstrated substantial success in video recognition tasks. The model was first pre-trained on the large-scale AVA action recognition dataset~\cite{gu2018ava}. The pre-trained model was then used for transfer learning on our dataset with the trimmed video clips of the driving action scenarios. 
%We train the models using both the RGB and the optical flow modalities separately. It was observed that the model trained with the optical flow modality performed better than the one trained with RGB data. Therefore, the flow-based TSN model was used to extract the video features for the explanation prediction task. 
This is illustrated as the Driving Behavior Recognizer module in Figure~\ref{figg:ModelArchitecture}, where the video backbone corresponds to the TSN model's Resnet backbone with optical flow frames as input and the recognition head corresponds to the TSN's average consensus head module for action prediction. The optical flow frames are extracted using the TVL1 algorithm \cite{zach2007duality} implemented in OpenCV.

\textbf{Important Object Track Explanation}: We use the ground truth important object tracks along with their explanation labels for training this task. The track bounding boxes were resampled at the frame numbers corresponding to the extracted video features from the flow-based TSN model. The resampled bounding boxes were then used to extract the per-object track video features by applying the Track-Of-Interest Align (TOI-Align) operator~\cite{singh2023spatio} to the video features. 
%The TOI-Align operator~\cite{singh2023spatio} is basically a combination of multiple ROI-Align operations applied to video frames using the corresponding track’s bounding boxes. The temporal dimension of the per-object track video features was reduced by simple averaging to obtain a more compact representation. 
Apart from this, the global driving action context information was also extracted by the spatial and temporal averaging of the video features from the flow-based TSN model. Finally, the global context information and the averaged per-track video features were concatenated and fed as input to a multi-layer perception (MLP) for explanation prediction. This is demonstrated as the Important Object Explanation Generator module in Figure~\ref{figg:ModelArchitecture}, where it can be seen that multiple tracks video features are simultaneously extracted and fed to the classifier for explanation prediction.

%The global context information helped improve the performance over using just the per-track features.

% add precision column
% our experiments show that consideration of heterogeneous road object categories help improve the identification task.
% add a row for class-conditioned bigru at last T.

% \textbf{Model} & 

\section{EXPERIMENTS AND RESULTS}

We split the IDD-X dataset on the basis of driving action scenarios into train 70\%, validation 15\%, and test 15\% sets. The number of ground truth important object annotations present in these driving scenarios comprised 6697 in training, 1728 in validation, and 264 in test. We consider the videos from the front view for the training and evaluation of all the tasks. 

%\subsection{Data Pre-processing, Training, and Evaluation Details}

\textbf{Important Object Track Identification}:
The dataset for this task was prepared using the YOLOv4 detection model~\cite{bochkovskiy2020yolov4} and the SORT tracking algorithm~\cite{bewley2016simple}. The YOLOv4 model was pre-trained on the COCO dataset \cite{lin2014microsoft} and
fine-tuned on the IDD dataset \cite{varma2019idd} for detecting road objects specific to Indian driving scenarios. The tracks for each driving scenario were obtained from the SORT tracker algorithm. A standard threshold of 0.5 was used for the average spatial IOU value between the predicted and the ground-truth object tracks. A total of 10882 predicted important object tracks and 108042 unimportant object tracks were obtained. 
% The difference in the ground truth and the predicted number of important object tracks is attributed to the failures of the tracking algorithm where in most cases a single object has multiple track predictions at different time intervals (or due to missed associations resulting in multiple track predictions for the same object). 

This data was used for training the BiGRU model with a hidden size of 5, and a single hidden layer. Given the highly skewed nature between the important and the unimportant object classes, we adopted mini-batch class-balanced sampling strategy while training this model with standard cross-entropy loss. A batch size of 32 and a learning rate of 0.001 with Adam optimizer \cite{kingma2014adam} were used for training the BiGRU model until 100 epochs.  Table~\ref{IOI_results} shows the comparative performances of our model with different inputs. Utilizing the object class information at all timesteps enabled the best performance of 86.8\% Recall and  50.7\% F1-Score. Our experiments show that consideration of heterogeneous road object categories throughout the temporal domain enriches the model's identification capability.

%The important object's Precision, Recall, and F1-score are reported for the evaluation of this task.
% used as evaluation metrics for the identification task. 

% The predicted track’s bounding box coordinates and the encoded object category were concatenated and fed as input to the BiGRU at every time step. 

%Table~\ref{IOI_results} shows the comparative performances of our model with different inputs. When the model was trained solely using the object tracks' bounding boxes, it achieved an F1-Score of 49.8\%. Incorporating object class information at the initial time step (T=0) had a slight drop in the F1-Score (49.6\%). The performance was nearly the same when the object class was conditioned at the last time step. When the model utilized object class information at all time steps, it achieved the highest Recall of 86.8\% and an improved F1-Score of 50.7\%.
%The quantitative results suggest that using the object tracks and the object classes across all time steps yields the best performance. 
% This aligns with our initial hypothesis that accounting for both spatial and categorical data throughout the temporal domain would enrich the model's identification capability.

\textbf{Driving Behavior Recognition}:
The input to the TSN model was randomly cropped video frames resized to 224x224, with random temporal sampling at the frame interval of 1, clip length 5, and number of clips 8.
 % extracted from the RGB data and then
The TSN model was trained with the standard cross-entropy loss, learning rate 0.000125, batch size 6, and for 40 epochs. The per-class recall and the weighted average accuracy of the predicted driving behaviors were used as evaluation metrics for this task. 
% Also, since the number of driving scenarios for slow down while left/right/U-turn categories were less, they were merged to a single class as slowing down while turning.
Table~\ref{ar_results} shows the performances of the TSN driving action recognition models using different modalities: RGB and optical flow. The model trained with optical flow modality shows significantly better performance with a 7.3\% improvement in overall accuracy. For specific driving actions, the per-class recalls also show superior performances with 1.9\%, 11.4\%, and 24\% improvements in 'Slowdown', 'Deviate', and 'Turn and Slowdown' classes respectively.
%The quantitative metrics indicate that the TSN model with optical flow modality outperforms its RGB counterpart in every evaluated category. 
This finding is crucial and substantiates our choice to employ the optical flow-based TSN model for feature extraction in the explanation prediction task.

\textbf{Important Object Track Explanation}:
The per-track video features were extracted using the trained TSN model with 256x224-sized image frames as input instead of 224x224. The frame width is modified so as to maintain the same aspect ratio with respect to the original image after resizing. This ensured unaltered aspect ratios for the track's bounding boxes as well. The video features extracted from the last layer of the Resnet backbone in the TSN model with spatial dimension 7x13, temporal size 8, and channel size 2048. The track bounding boxes were downsampled to the video feature spatial dimensions before the application of the TOI-Align operator. The output from TOI-Align was 1x1x2048x8 (width x height x channel x time) per-object video features. The global context information obtained from spatial and temporal averaging of the video features was of dimension 1x1x2048x1. The concatenated output, using the global context information and the per-object track features, was of dimension 4096. This was fed as input to an MLP with a hidden layer of size 128. The explanation labels in the training dataset with a total count of less than 100 were not considered for the training and evaluation of this task. Standard cross-entropy loss with learning rate 0.001, and batch size 16, was used for training the MLP classifier until epochs 50. Per-class F1-score, average F1-score, and weighted average F1-score are reported as evaluation metrics for this task. 

We compare two variants: TOI-Aligned per-track features, and TOI-Aligned per-track features combined with global context information. As observed from the Table~\ref{explanation_results}, the TOI-Aligned model excels in the 'Congestion', 'Avoid Congestion', and 'Overtaking' categories, with better per-class F1 scores. When the global context information is added, the performance of the tail classes ('Crossing', 'Interfering Being', 'Cut-in', and 'Avoid On-road Being') improves by a large margin (7.1\% for average F1-score, 1\% for weighted average F1-score). Overall, the results suggest that inclusion of global context information significantly improves the model's ability to correctly identify the underprivileged explanation categories. However, for more common driving situations (e.g. `Congestion' and `Overtaking') per-track-features-only appear to suffice. The performance in categories such as `Cut-in' and `Avoid On-road Animal' is low across both variants, pointing out areas for improvement.

% Overall, the quantitative results indicate the efficacy of using the per-track features and the global video context features in explaining the important objects in diverse driving scenarios. Future work will focus on optimizing these feature combinations and exploring other architectures to improve performance further.

% \section{CONCLUSIONS}

% We propose a new problem of ego-relative important object localization and explanation in dense and unstructured traffic. To address this problem, we collected a novel multi-view driving video dataset, IDD-X. The dataset contains ego-relative annotations for multiple important road objects with their spatial and temporal locations, explanations, and class labels in video clips. Additionally, the ego vehicle's driving behavior is provided for every video clip to study the influence of multiple important objects on its driving behavior. Leveraging the dataset capabilities, we introduce custom-designed deep network architectures for multiple important object localization and per-object explanation prediction. Our preliminary results show that the proposed localization and explanation tasks are challenging. Better architectures and approaches could be explored in the future to further improve the performance for importance and explanation prediction of heterogeneous road objects in complex driving scenarios.

% \section*{ACKNOWLEDGMENT}

% We thank iHubData, the Technology Innovation Hub (TIH) at IIIT-Hyderabad for supporting this project.

\textbf{Acknowledgement} The project is funded by the iHubData and Mobility at IIIT Hyderabad. We thank the data collection and annotation team for their effort.

% \section{CONCLUSIONS}

% A conclusion section is not required. Although a conclusion may review the main points of the paper, do not replicate the abstract as the conclusion. A conclusion might elaborate on the importance of the work or suggest applications and extensions. 

%                                   % the textheight of the last page by a suitable amount.
%                                   % This command does not take effect until the next page
%                                   % so it should come on the page before the last. Make
%                                   % sure that you do not shorten the textheight too much.

%%%%%%%%%%%%%%%%%%%%%%%%%%%%%%%%%%%%%%%%%%%%%%%%%%%%%%%%%%%%%%%%%%%%%%%%%%%%%%%%

%%%%%%%%%%%%%%%%%%%%%%%%%%%%%%%%%%%%%%%%%%%%%%%%%%%%%%%%%%%%%%%%%%%%%%%%%%%%%%%%



% \section*{ACKNOWLEDGMENT}

% The preferred spelling of the word ÒacknowledgmentÓ in America is without an ÒeÓ after the ÒgÓ. Avoid the stilted expression, ÒOne of us (R. B. G.) thanks . . .Ó  Instead, try ÒR. B. G. thanksÓ. Put sponsor acknowledgments in the unnumbered footnote on the first page.

%%%%%%%%%%%%%%%%%%%%%%%%%%%%%%%%%%%%%%%%%%%%%%%%%%%%%%%%%%%%%%%%%%%%%%%%%%%%%%%%

\bibliographystyle{IEEEtran}
\bibliography{IEEEfull}

\end{document}